\documentclass[10pt,twocolumn,letterpaper]{article}

\usepackage[pagenumbers]{wacv} 

\usepackage{makecell}

\usepackage{algorithm}
\usepackage[noend]{algpseudocode} 
\usepackage{tikz}
\usepackage{float}
\usepackage{svg}

\definecolor{wacvblue}{rgb}{0.21,0.49,0.74}
\usepackage[pagebackref,breaklinks,colorlinks,allcolors=wacvblue]{hyperref}

\def\wacvPaperID{1508} 
\def\confName{WACV}
\def\confYear{2027}

\title{Scalable Detection of Fossil Palynomorphs in Multifocal Digital Microscopy Images}

\author{Abbas Shaikh\\
Rice University\\
Houston TX USA\\
{\tt\small abbas@alumni.rice.edu}
\and
Praise Mayor\\
Rice University\\
Houston TX USA\\
{\tt\small pom1@rice.edu}
\and
Patrick Ainlay-Vazquez\\
Rice University\\
Houston TX USA\\
{\tt\small pta2@rice.edu}
\and
Aditya Viswanathan\\
Rice University\\
Houston TX USA\\
{\tt\small av77@rice.edu}
\and
Teon Golden\\
Rice University\\
Houston TX USA\\
{\tt\small tjg7@rice.edu}
\and
Eric Zhang\\
Rice University\\
Houston TX USA\\
{\tt\small erictz@alumni.rice.edu}
\and
Ingrid C. Romero\\
Smithsonian National Museum of Natural History\\
Washington DC USA\\
{\tt\small RomeroIC@si.edu}
\and
Alexander E. White\\
Smithsonian Office of Digital and Innovation\\
Washington DC USA\\
{\tt\small WhiteAE@si.edu}
\and
Scott Wing\\
Smithsonian National Museum of Natural History\\
Washington DC USA\\
{\tt\small wings@si.edu}
\and
Arko Barman\\
Rice University\\
Houston TX USA\\
{\tt\small arko.barman@rice.edu}
}

\begin{document}
\maketitle
\begin{abstract}
Palynomorphs (microscopic, organic-walled fossils such as pollen, spores, and dinoflagellates) are important high-resolution records of past climates and are critical to the study of ancient ecosystems. Existing methods rely on manual analysis of high-resolution, multifocal digital microscopy images, which is slow and time-consuming and requires researchers to compromise on the scale of their investigations. To the best of our knowledge, our work proposes the first ever scalable end-to-end pipeline for automated palynomorph detection in whole slide images that addresses this bottleneck through: (1) efficient methods for decomposing and compressing digitized multifocal microscope slide images into tractable 2-dimensional tiles for analysis; (2) benchmarking modern object detection models, including RF-DETR, for the detection of palynomorphs, achieving an AP@50 of 0.879; (3) an efficient algorithm for the synthesis of detection outputs across large-scale, high-resolution images; and (4) an I/O optimization resulting in faster inference time. Our methods drastically reduce the time required for palynomorph detection in a single slide from often days of manual inspection to under one hour of automated analysis, enabling palynological research at a substantially greater scale.
\end{abstract}
    
\section{Introduction}
\label{sec:intro}

Palynomorphs -- microscopic, organic-walled fossil particles including pollen, spores, dinoflagellates, fungi, and algae -- are among the most abundant and best-preserved plant fossils in the geological record~\cite{traverse2007paleopalynology, zargar2025pollen}. The study of palynomorphs is of critical importance in paleoclimatological research, acting as direct and resilient records of past environments that inform reconstructions of Earth's past climates and projections of future ones~\cite{zargar2025pollen, egan_historical_2005}. In particular, palynomorphs from ancient warm intervals, when CO\textsubscript{2} was elevated, offer empirical analogues for projecting ecosystem response under continued warming~\cite{burke2018pliocene}.

The Smithsonian National Museum of Natural History 
stewards a collection of tens of thousands of microscope slides of palynomorphs, in which a single slide can contain thousands of specimens~\cite{romero_digitizing_2026}. While many slides have been digitized, only a small fraction have been actively studied. Manual analysis is often slow and time-consuming, as palynomorphs are interspersed with other sediments, making them difficult to distinguish and localize quickly. However, digitizing these slide collections opens the possibility of using machine-learning object detection models to support scalable, high-throughput palynological research~\cite{punyasena2022automated, romero_digitizing_2026, martinsen_3-billion_2024, martinsen_fossil_2026, jaramillo_digitizing_2026}.

Digitized slides are scanned at several magnifications and focal planes, producing several high-resolution images often exceeding $25$\,GB in total ~\cite{romero_digitizing_2026}. Given the sheer volume of data in a single digitized slide, processing an image in its entirety is computationally infeasible. This challenge prohibits the use of standard techniques for pre-processing, modeling, and post-processing used in standard object detection frameworks. These challenges are further compounded by the fact that multifocal images capture significantly more information and complexity than a standard two-dimensional image. 

\begin{figure}[t]
    \centering
    \begin{subfigure}[t]{0.66\linewidth}
        \centering
        \includegraphics[width=\linewidth]{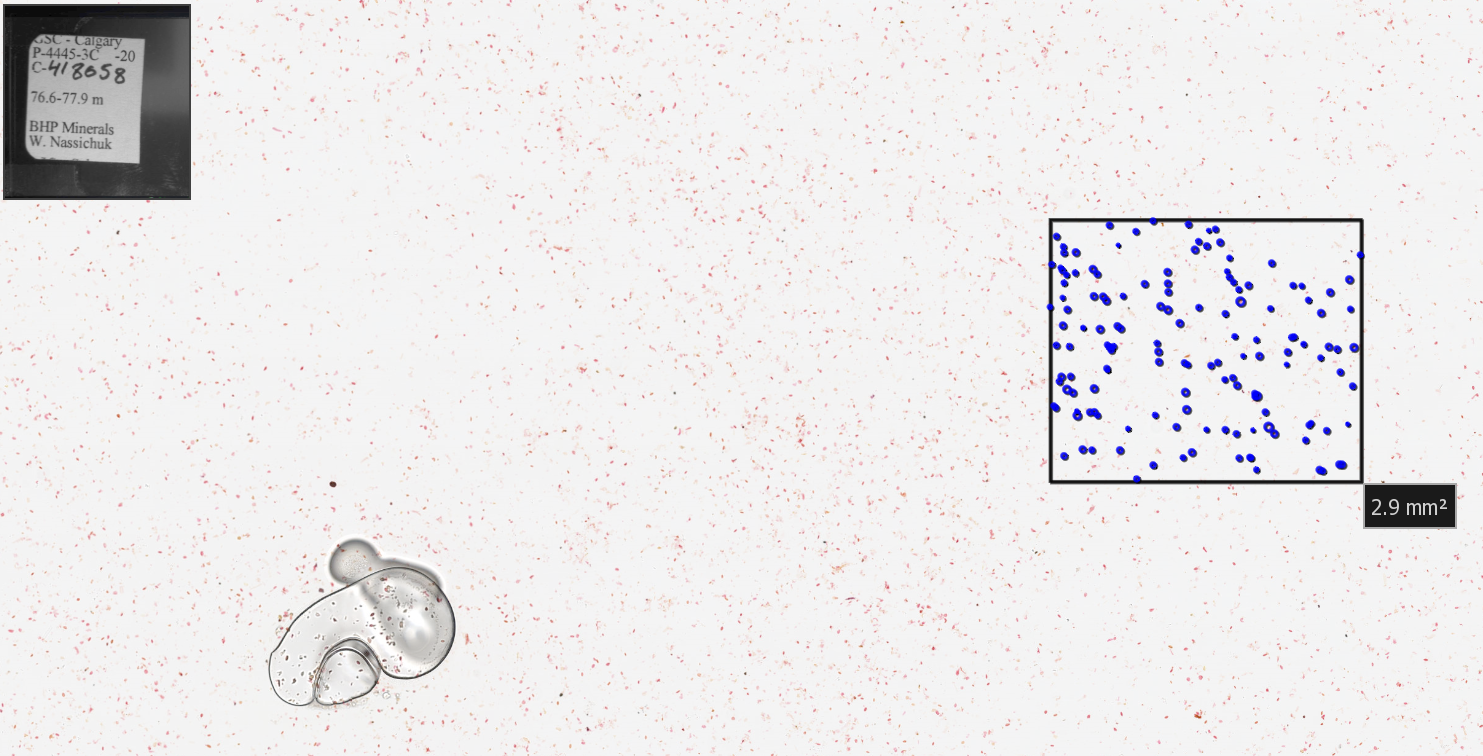}
        \caption{High-level crop of a slide image with annotations.}
        \label{fig:crop_a}
    \end{subfigure}
    \begin{subfigure}[t]{0.33\linewidth}
        \centering
        \includegraphics[width=\linewidth]{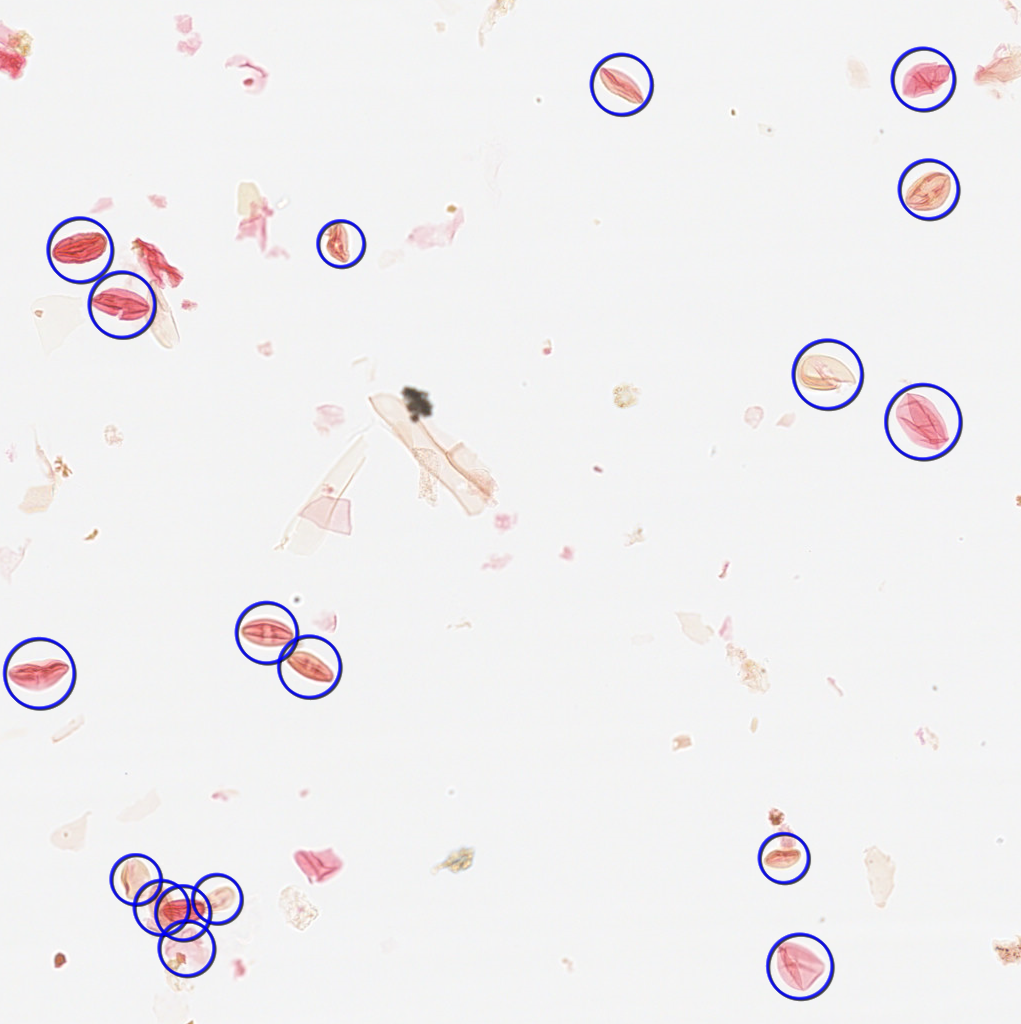}
        \caption{Crop at 10$\times$ magnification with palynomorphs circled.}
        \label{fig:crop_b}
    \end{subfigure}
    \caption{Sample slide image crops with human annotations at different scales.}
    \label{fig:crop}
\end{figure}

Each palynomorph specimen is a three-dimensional body suspended in mounting medium, and no single focal plane in a slide brings every specimen into focus. Specimens are also frequently dense, overlapping, or occluded by debris of similar contrast and texture. Furthermore, variations in staining, mounting medium, and acquisition parameters induce substantial appearance shifts across slides. Hence, a model fit to one preparation cannot necessarily be assumed to transfer to another. Figure~\ref{fig:crop} shows a representative annotated crop illustrating the density, variation in scale, and background clutter that characterize the data. 

Due to these challenges, automated palynology has largely focused on classification of pre-cropped specimens, whether from single images~\cite{kong2016spatially, punyasena2022automated} or by treating image stacks as 3D sequential data~\cite{daood2016classifying, romero2020improving}. Detection has been addressed more recently~\cite{gallardo2019precise, kubera2022detection, long2025hieraedgenet, shi2026pollen, biricz2025efficient, dhawan2026two}, but on fields of view of a few megapixels only~\cite{gallardo2019precise}, not on whole slide images. 

The challenges of decomposing high-resolution, multifocal data from whole slide images remain unaddressed; hence, the throughput problem that motivates automation remains unsolved. Multifocal information is also seldom used for detection, the most common approach being to apply an object detection method once per plane and fuse the resulting proposals across planes with non-maximum suppression (NMS)~\cite{gallardo2019precise}. This strategy is impractical for most palynological research: for multifocal images consisting of 25--27 focal planes and up to 10,000 tiles per slide, detection and aggregation become exponentially costly. To the best of our knowledge, no prior work performs palynomorph detection at a whole-slide scale, and none reports an end-to-end system capable of performing detection on a complete, multifocal digitized slide. 

Moreover, our end-to-end pipeline performs inference on entire slides at a  pace comparable with the digitization of slides. While our baseline methods yield a runtime of roughly 8--9 hours per slide, our I/O optimization strategy reduces inference time to less than 1 hour per slide. Additionally, to the best of our knowledge, our work is the first to detect all organic remains that can fall into the palynomorph category, not focusing solely on pollen and spores, as seen in all prior research~\cite{gallardo2019precise, kubera2022detection, long2025hieraedgenet, shi2026pollen, biricz2025efficient, dhawan2026two}.


\noindent Our contributions can be summarized as follows:
\begin{itemize}
    \item We propose an end-to-end whole-slide palynomorph detection pipeline leveraging state-of-the-art object detection models, including You Only Look Once 26 (YOLO26)~\cite{jocher2026ultralyticsyolo26unifiedrealtime} and Roboflow DEtection TRansformer (RF-DETR)~\cite{robinson2025rf}, usable at operational scale with inference times comparable to the digitization of microscope slide images.
    \item We develop a scalable preprocessing pipeline for high-resolution, multifocal microscopy data that crops images into tiles that are tractable for processing and compresses each focal stack into a single 2D image, yielding inputs directly usable by object detection models.
    \item We introduce a \emph{boundary-aware} variant of the non-maximum suppression procedure to merge and de-duplicate detections across thousands of tiles efficiently.
    \item We devise an I/O optimization method to speed up inference about 10 times, resulting in an inference time of less than 1 hour per slide.
\end{itemize}
\section{Related Work}
\label{sec:related_work}

The development of AI and computer vision methods is essential in paleontology and palynology to address limitations in manual techniques, including subjectivity and efficiency \cite{bhoyar2025revolutionizing}. Historically, computational analysis of palynomorphs has primarily focused on the classification of pre-cropped specimen images into different species. For instance, several methods have been proposed for purely classification problems, such as spatially aware dictionary learning for species-level recognition of fossil pollen~\cite{kong2016spatially} and convolutional neural networks (CNNs) for taxonomic classification of pollen~\cite{punyasena2022automated}. Similarly, machine learning methods have been proposed to refine fossil pollen taxonomy using super-resolution microscopy~\cite{romero2020improving}. 

More recently, CNN-based object detection methods have been applied to automate localization. Previous work has explored a variety of methods for object detection, including You Only Look Once (YOLO), Faster Region-based CNN (Faster R-CNN), and RetinaNet, for both localization and recognition of pollen in microscopy data~\cite{gallardo2019precise, kubera2022detection}. There have been methodological advances in CNN-based pollen detection, such as HieraEdgeNet~\cite{long2025hieraedgenet}, which incorporates multi-scale edge-enhanced features into model reasoning to address the challenging visual features of microscopic specimens, such as indistinct borders and highly complex background debris. Pollen-YOLO, a pollen detection method based on YOLO, leverages an improved channel–spatial collaborative attention module to ensure robustness to morphological variability and distortion of specimens within images~\cite{shi2026pollen}.

While transformer-based architectures have now emerged as powerful alternatives for object detection, their use in palynology remains relatively unexplored. Recent work has investigated the use of pre-trained vision transformers via linear probing as robust feature extractors in hybrid CNN-transformer architectures for automated pollen detection \cite{biricz2025efficient}. Real-Time DEtection TRansformer (RT-DETR) has also been applied as a localization tool within a two-stage semi-supervised framework for the identification and classification of pollen particles in evanescent wave scattering microscopy~\cite{dhawan2026two}. In addition, while existing deep learning frameworks largely process specimens as isolated 2D planes, prior work has highlighted the value of incorporating multifocal input to resolve structural occlusion and blur~\cite{daood2016classifying}. Thus, there remains an unaddressed need to leverage modern transformer architectures for scalable end-to-end object detection that leverages multifocal inputs in palynology and paleontology.

\section{Data}
\label{sec:data}

The dataset used in this study consists of 847 microscopy slides provided by the Smithsonian National Museum of Natural History, collected from sites across North America. Each slide was digitized using a NanoZoomer Slide Scanner, which outputs data in the NanoZoomer Digital Pathology Image (NDPI) format. Each resulting image is a scanned area of $20$mm $\times$ $20$mm from a slide at varying magnifications that can contain tens of thousands of palynomorphs. At each magnification, the image captures a multifocal view across 25--27 focal planes. Due to the high resolution of digital microscopy images, each slide can be greater than 25 gigabytes in size~\cite{romero_digitizing_2026}.

Among these 847 slides, 82 slides were selected for annotation and curation for this study. Each such image is paired with a NanoZoomer Digital Pathology Annotation (NDPA) file, an XML-based file that stores annotations, measurements, and regions of interest. For each slide, one or more rectangular regions of interest of 2.9mm$^2$ were selected and manually annotated with all palynomorphs by researchers at the Smithsonian. 

The images within the dataset capture a wide range of visual characteristics, including multiple staining methods and mounting media, various densities and distributions of palynomorphs (as well as different palynomorph subtypes), along with both organic and inorganic matter. These variations, collected from diverse environments, all produce variability that informed our modeling choices.

Figure~\ref{fig:crop_a} presents a high-level crop of a slide image within the dataset, providing visual evidence of the diversity and scale of organic and inorganic matter found within a single slide image. The 2.9mm$^2$ annotated region of interest can be seen outlined in black, with individual palynomorphs outlined in blue. A closer view of the slide shown in Figure~\ref{fig:crop_b} displays a smaller region containing identified palynomorphs at higher resolution. As shown, the volume and diversity of data within a given slide underscores the need for automated methods of analysis.

\begin{figure}[t]
    \centering
    \includegraphics[width=0.33\linewidth]{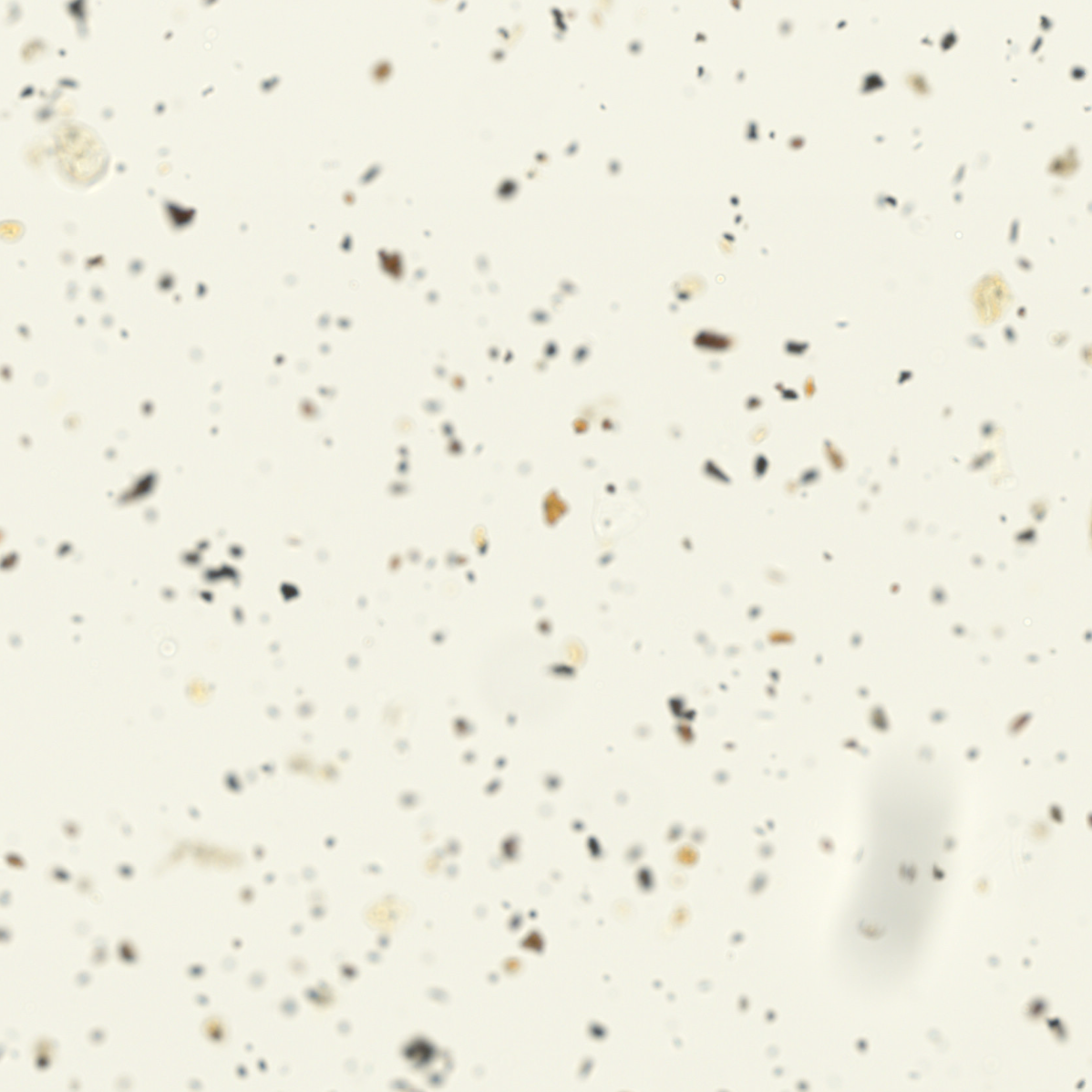}        
    \includegraphics[width=0.33\linewidth]{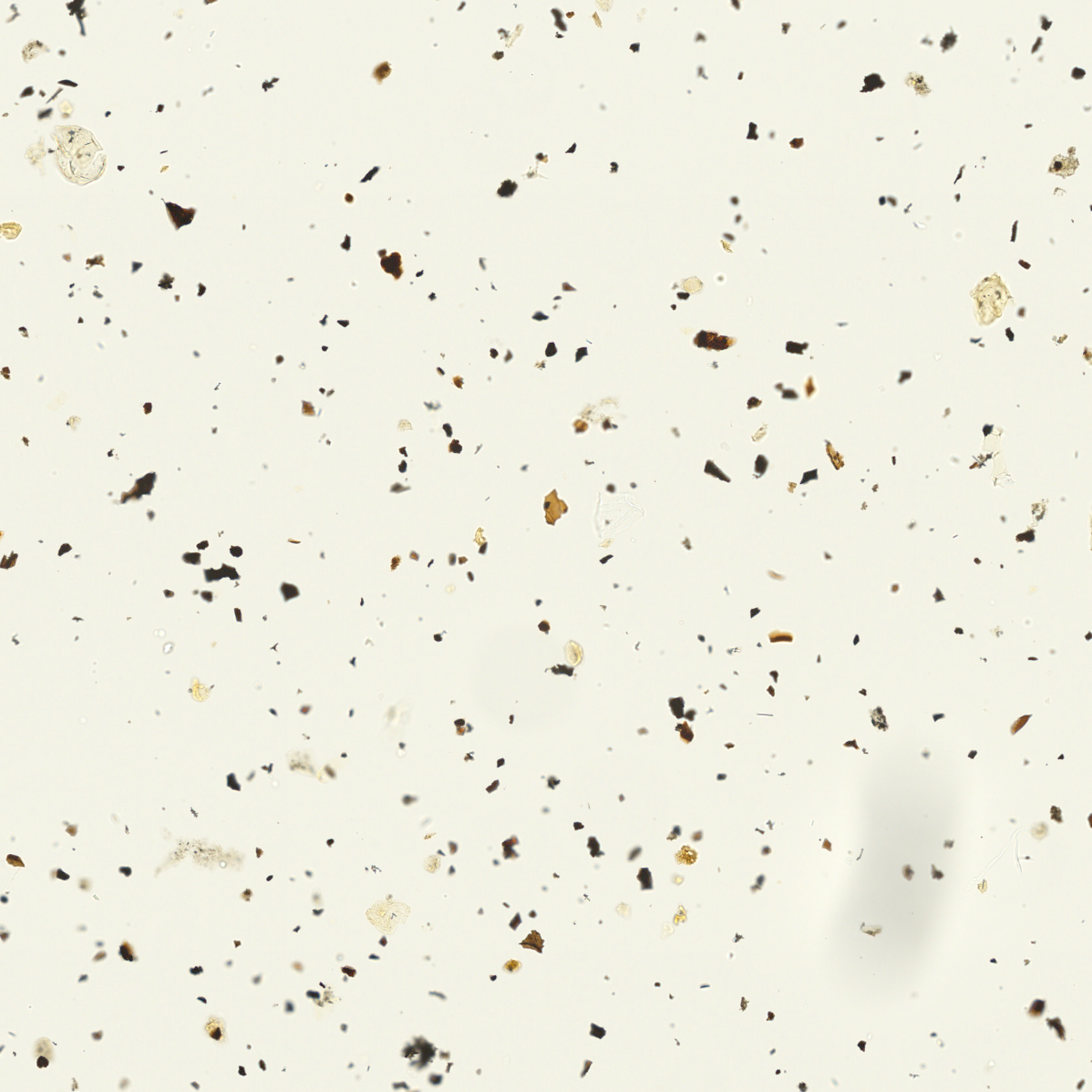}
    \caption{Crops at 15$\times$ magnification from the $-12 \mu m$ (left) and $0 \mu m$ (right) focal planes of a slide.}
    \label{fig:focal-planes}
\end{figure}

Figure \ref{fig:focal-planes} shows $2$ different focal planes ($z = -12 \mu m$ and $z=0 \mu m$) of the same region of a slide at $15\times$ magnification. The slide was originally imaged at $25$ focal planes, encompassing all positions in the z-stack from $-12 \mu m$ to $12 \mu m$. It can be observed that the sharpness of the image varies considerably across focal planes, and in this example, the $0 \mu m$ focal plane provides the clearest image. 

We note that the objects on the slides are suspended in a medium and free to move in a 3D space. As a result, individual structures of a single palynomorph may be located in different focal planes, and different objects on the same slide may be in focus in different focal planes. As a result, a single globally optimal focal plane (where all objects appear sharply visible) typically does not exist for any multifocal slide images of this nature. Thus, considerable pre-processing is needed to determine how to effectively compress information across multiple focal planes for efficient 2D object detection.
\section{Methods}
\label{sec:methods}

\subsection{Overview}

\begin{figure*}[ht]
\centering
\includegraphics[trim = 0 35 0 35, clip, width=0.7\textwidth]{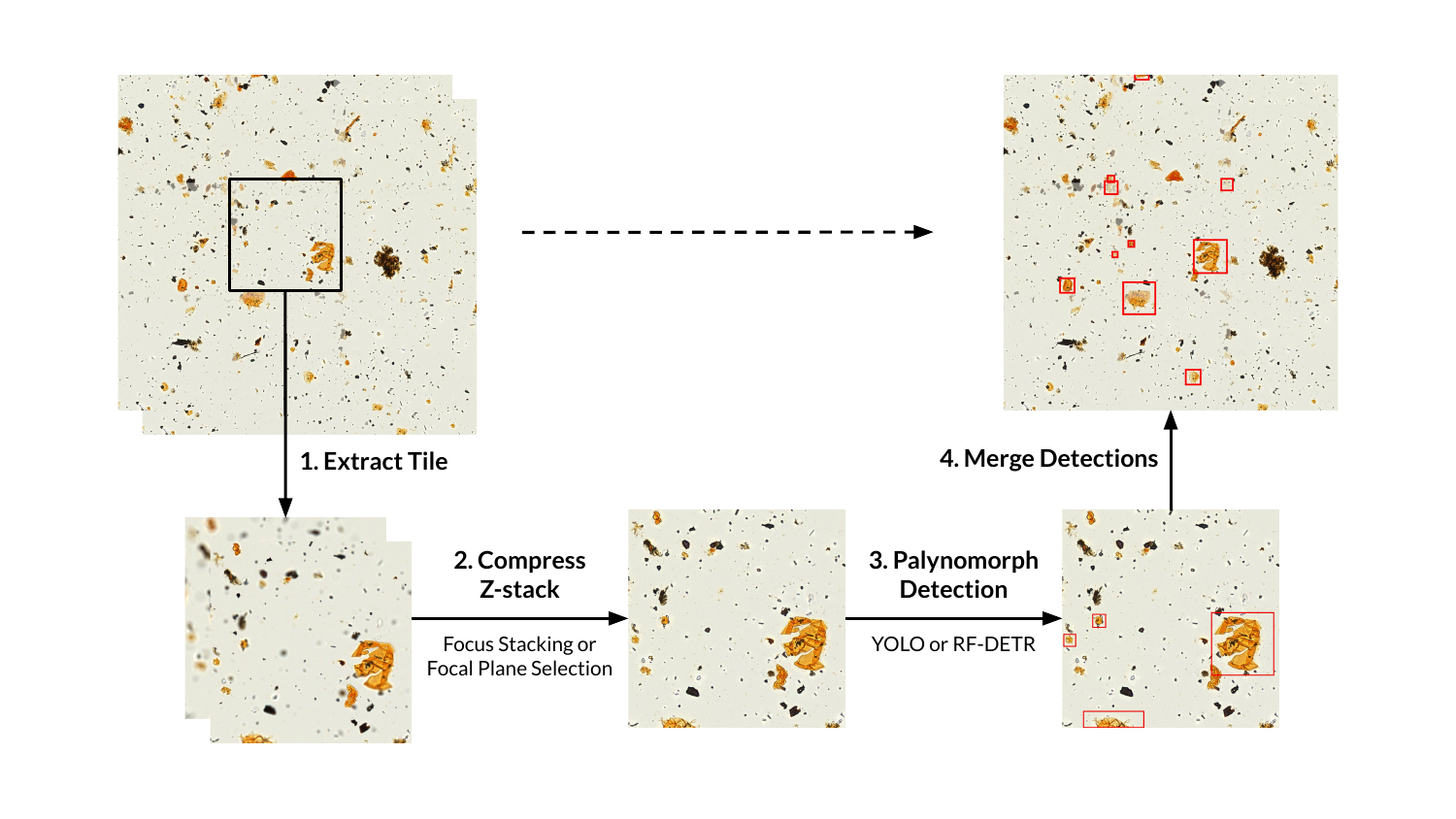}
\caption{Our end-to-end palynomorph detection pipeline for digitized multifocal whole slide images.}
\label{fig:pipeline}
\end{figure*}

Processing of high-resolution NDPI images is time and resource-intensive. We introduce an efficient, end-to-end pipeline to split an image into manageable tiles, convert these multifocal tiles into compressed, two-dimensional representations that can be used to train and test object detection models, and finally synthesize tile-level predictions to produce complete palynomorph detection results on whole slides. We explored two methods for compression of multifocal images: (1) focus stacking, which combines information across focal planes; and (2) focal plane selection, which identifies the sharpest individual plane. These two techniques were used in conjunction with significant data augmentation to train and evaluate two state-of-the-art object detection models: YOLO26~\cite{jocher2026ultralyticsyolo26unifiedrealtime} and RF-DETR~\cite{robinson2025rf}. Finally, we integrated the trained models into a scalable pipeline with optimized tile reading, focal plane compression, and an efficient post-processing method to aggregate and de-duplicate tile-level predictions across whole-slide images. These methods enable efficient and accurate automated analysis of high-resolution, multifocal microscopy images of palynomorphs. Figure~\ref{fig:pipeline} shows our end-to-end pipeline for palynomorph detection for whole-slide images.

\subsection{Pre-processing}
\label{sec:preprocessing}

As noted, processing raw NDPI files in their entirety is computationally infeasible given the memory constraints of standard hardware. Consequently, the following pre-processing pipeline was used to extract and organize relevant data into a usable format. In particular, the pipeline serves to (1) efficiently decompose large images into smaller tiles and (2) compress all focal planes for each multifocal image into a single 2-dimensional image, which can be readily used for training and evaluating state-of-the-art object detection models. 

For model development, we select only the 2.9mm$^2$ regions of interest containing ground truth annotated palynomorphs from each image. First, we subdivide these large annotated regions into fixed-size tiles (e.g., $1024 \times 1024$ pixels) at the maximum magnification of $40\times$ and an overlap of 10\%. This approach reduces the input size for the model while maintaining high resolution and sufficient scale to contain larger specimens. An overlap of 10\% was empirically chosen to ensure that palynomorphs spanning tile boundaries were fully captured within at least one tile, enabling effective compositing of detections from adjacent tiles during post-processing. 

Ground truth annotations, stored in physical nanometer coordinates relative to the slide center, were mapped onto the image pixel grid at the specified magnification using an affine transformation. Additionally, while the source data contains circular ground truth annotations, they were transformed to standard rectangular bounding box coordinates $(x, y, w, h)$ that bound these circular regions, where $(x, y)$ indicates the the coordinates of the center of the bounding box for YOLO26 or the coordinates of the top left corner for RF-DETR, $w$ is the width, and $h$ is the height of the bounding box. In addition, bounding boxes were excluded from a given tile if their intersection with the tile boundary retained less than 25\% of the ground truth annotated area, thereby preventing the inclusion of marginal object fragments as positive training samples. Extracted tiles were stored as 4-dimensional arrays with shape $(h, w, c, z)$, where $h$ is height, $w$ is width, $c$ is the number of channels, and $z$ is the number of focal planes. 

We employ two approaches to compress information from each multifocal image into a single 2D representation:

\begin{enumerate} 
\item \textbf{Focus Stacking:} Since important information about structures within the image may be present across multiple focal planes, we employ a technique called focus stacking to combine information from all planes into a single 2D image. In particular, the Laplacian of Gaussian (LoG) was used as a focus measure operator and applied to each focal plane. The LoG operator first smooths an image with a $5\times5$ isotropic Gaussian filter to reduce noise, and then applies a $5\times5$ isotropic Laplacian filter to detect regions of rapid intensity change (edges and fine detail), which are most pronounced in well-focused images. Each pixel in the output is drawn from whichever focal plane produced the maximum LoG response at that position.

\item \textbf{Focal Plane Selection:} Our other method selects the single most in-focus plane without modifying its contents using the Tenengrad score, rather than compressing information across planes. Tenengrad score is computed as the sum of squared Sobel gradients. As a result, a peak in the Tenengrad score is observed in the plane with the strongest, most well-defined edges. Thus, the plane with the highest Tenengrad score was chosen as the representative plane for each multifocal tile.
\end{enumerate}

\begin{figure}[t]
    \centering
    \begin{subfigure}[t]{0.49\linewidth}
        \centering
        \includegraphics[width=\linewidth]{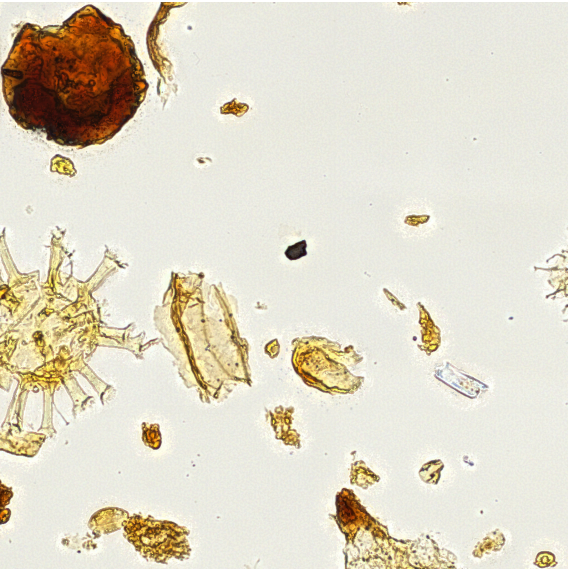}
        \caption{A focus-stacked tile.}
        \label{fig:focus-stacking_a}
    \end{subfigure}
    \begin{subfigure}[t]{0.49\linewidth}
        \centering
        \includegraphics[width=\linewidth]{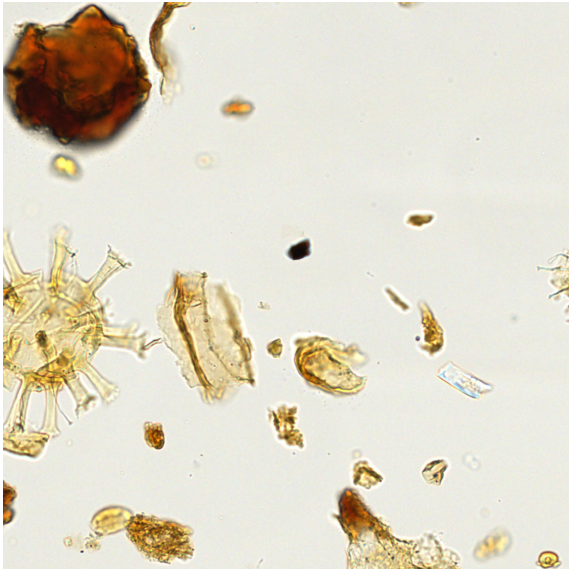}
        \caption{The representative plane of a tile.}
        \label{fig:focus-stacking_b}
    \end{subfigure}
    \caption{A side-by-side comparison of focus stacking and focal plane selection.}
    \label{fig:focus-stacking}
\end{figure}

A comparison between a focus-stacked tile and the representative plane of the tile is shown in Figure \ref{fig:focus-stacking}. Notably, while a singular focal plane preserves most of the image in focus, larger structures tend to have some components that still remain out of focus. The focus-stacking procedure leverages information from other planes where these structures might be in focus to produce a higher contrast image. However, the trade-offs of using focus-stacking include increased noise and visual artifacts. Since adjacent pixels may be sourced from different planes, the stacking process can introduce discontinuities that are not edges or corners.

\subsection{Models}
\label{sec:models}

The datasets of extracted tiles for both focus-stacking and focal plane selection were used to train and evaluate two object detection models. We frame the palynomorph detection as a single-class object detection task. To evaluate the effectiveness of modern object detection architectures for this task, we benchmark state-of-the-art convolutional neural networks and transformer-based models. In particular, we adopt the YOLO26-L architecture \cite{jocher2026ultralyticsyolo26unifiedrealtime} as our CNN baseline along with RF-DETR-2XL \cite{robinson2025rf}, which currently achieves state-of-the-art performance in object detection on standard COCO benchmarks~\cite{lin_microsoft_2014}.

To mitigate the sensitivity of the models to variations in staining, mounting media, image acquisition parameters, and ecological diversity present within the Smithsonian's palynological collection, we incorporate extensive data augmentation to improve the robustness of our pipeline. Our augmentation includes affine transformations; flips; hue, saturation, and value shifts; as well as mosaic augmentation.

\subsection{Whole-Slide Inference}
\label{sec:whole-slide}

Obtaining palynomorph detection inference in a full slide poses additional computational challenges compared to training a model on pre-processed tiles. Namely, it requires reading each tile and compressing its focal planes in direct sequence with detection across thousands of tiles. Furthermore, tile-level detections must then be synthesized into a single cohesive set of detections across a whole slide. To handle the computational constraints of this task, we employ an additional I/O optimization and an efficient post-processing algorithm described below.

\subsubsection{I/O Optimization}
\label{sec:io}
In initial experiments, we noted GPU utilization near 0\% for extended periods, with per-tile inference runtime dominated by reading the NDPI file rather than by the detector model. Hence, we designed three changes to the slide input path for I/O optimization:
\begin{enumerate}
\item \textbf{Handle Caching:} NDPI files store each focal plane as a sequence of JPEG-compressed horizontal strips, each spanning the full width of the slide. We read the NDPI file handle and the strip index for each focal plane only once per slide and reuse them, rather than reconstructing on every tile read across all focal planes.
\item \textbf{Band Reading:} A single tile covers only a small horizontal fraction of the strips it intersects, so reading tiles independently forces each strip to be decoded again for every tile along its width. As a result, over a region spanning many tiles in width, the same compressed data is decoded many times. To optimize these I/O reads, we instead read a full-width band of tiles in a single operation, so that each strip is decoded exactly once, and crop individual tiles from the band in memory.
\item \textbf{Parallel Input Preparation:} We run tile reading and focal plane compression in parallel CPU worker processes that feed the GPU, overlapping input preparation of subsequent tiles (parallelizing both I/O reads and focal plane compression) with inference on the current tiles. 
\end{enumerate}
We empirically verified that each optimization leaves the output of our end-to-end pipeline unchanged. Tiles sliced from bands are pixel-exact against individual per-tile reads, and the serial and parallel paths produce identical detections (bounding box coordinates and dimensions as well as confidence scores) on whole test slides. Together, these changes reduce input cost from approximately 3\,s to 0.2\,s per tile, and the end-to-end whole-slide inference time by roughly an order of magnitude (Section~\ref{sec:results}). 

\subsubsection{Post-processing}
\label{sec:postprocessing}
Each slide can span up to 100,000 pixels in each dimension, resulting in up to 10,000 tiles per slide image. At inference time, tiles are passed through the model independently, producing a set of bounding box predictions with associated confidence scores. Collecting predictions across all tiles yields a dense set of candidate annotations over the entire slide. Since adjacent tiles overlap, objects near tile boundaries may be detected multiple times across neighboring tiles, producing duplicate annotations. 

The most commonly employed solution to resolve duplicate detections is NMS, which greedily retains the highest-confidence annotation and suppresses any overlapping annotation whose Intersection-over-Union (IoU) with the retained box exceeds a fixed threshold, iterating until no further suppression occurs (\cite{neubeck2006efficient}). Applied naively across all tiles, this would require comparing every pair of annotations from all 10,000 tiles, which is often computationally infeasible. To enable efficient whole-slide annotation, we additionally introduce a boundary-aware variant of NMS described in Algorithm~\ref{alg:boundary-nms}.

For the $i$-th tile ($T_i$), the set of detections within $T_i$ is denoted as $\mathcal{D}_i = \{d = (b, s)\}$, where $b$ is the bounding box of a detection and $s$ its associated confidence score. For each pair of overlapping tiles, $(T_i, T_j)$, we compute the IoU between pairs of detections, $(d_a, d_b)$, where $d_a \in \mathcal{D}_i$ and $d_b \in \mathcal{D}_j$. All pairs exceeding a threshold $\delta$ are merged via a union-find structure $\mathcal{U}$. Across successive iterations, duplicate detections are grouped globally into connected components, and the highest-scoring detection within each component is efficiently retrieved via the find operation. By restricting de-duplication to detections that fall only within overlapping tiles and excluding detections in non-overlapping tiles, the algorithm avoids the quadratic comparison cost of traditional NMS, keeping whole-slide post-processing tractable. 

\begin{algorithm}
\caption{Boundary-Aware NMS for Whole-Slide De-duplication}
\label{alg:boundary-nms}
\begin{algorithmic}[1]

\Require Detections per tile $\{(T_i, \mathcal{D}_i)\}$, IoU threshold $\delta$

\State Sort tiles $T_1, \ldots, T_n$ in row-major order
\State Flatten all detections into global list $\mathcal{D}$, recording per-tile indices
\State Initialize union - find $\mathcal{U}$ over $|\mathcal{D}|$ elements

\For{each tile $T_i$}
    \For{each later tile $T_j$ overlapping $T_i$}
        \For{$d_a \in \mathcal{D}_i$, $d_b \in \mathcal{D}_j$}
            \If{$\mathrm{IoU}(d_a.b, d_b.b) > \delta$}
                \State $\mathcal{U}.\textsc{Union}(d_a, d_b)$
            \EndIf
        \EndFor
    \EndFor
\EndFor

\State $\mathcal{C} \gets \{\}$
\For{each $d \in \mathcal{D}$}
    \State $r \gets \mathcal{U}.\textsc{Find}(d)$
    \If{$r \notin \mathcal{C}$ \textbf{or} $d.\mathit{s} > \mathcal{C}[r].\mathit{s}$}
        \State $\mathcal{C}[r] \gets d$
    \EndIf
\EndFor

\State \Return $\{\mathcal{C}[r] : r \in \mathcal{C}\}$

\end{algorithmic}
\end{algorithm}

\section{Experiments and Results}
\label{sec:experiments}

\subsection{Experiment Design}
\label{sec:setup}

\textbf{Data Splits.} Of the 82 annotated slides described in Section~\ref{sec:data}, we select 58 for training, 12 for validation, and 12 as a held-out test set (a 70/15/15 split), yielding 4{,}585, 1{,}288, and 986 tiles in each split, respectively. While the number of tiles between splits is not uniform, splitting at the slide level prevents data leakage across splits, which is especially important since every slide may visually differ due to variations in staining, mounting medium, and specimen distribution. Additionally, we verified that annotation density, palynomorph subtype distribution, and staining method were comparable across the three splits prior to training.

\textbf{Training.} YOLO26-L~\cite{jocher2026ultralyticsyolo26unifiedrealtime} was trained on tiles at a resolution of $1024\times1024$, with a composite bounding-box regression and Complete IoU (CIoU) loss, applying early stopping with a patience of 10 epochs. RF-DETR-2XL~\cite{robinson2025rf} was trained on tiles resampled to a resolution of $1000\times1000$ (to accommodate the backbone's stride requirement) with a Hungarian-matching objective that combines $L_1$ and generalized IoU losses, under the same early-stopping criterion and smoothed using an exponential moving average of its weights. Both models were optimized with AdamW (batch size 16) on a single NVIDIA L40S GPU, and the final checkpoint for each run was selected based on the peak validation AP@50-95 (mean of average precision values across IoU thresholds from 0.5--0.95 in increments of 0.05). For each pair of focal plane compression (focus stacking vs. focal plane selection) and model, we tuned the optimizer, regularization, and augmentation hyperparameters with Optuna's Bayesian optimization framework~\cite{optuna_2019}, with the objective of maximizing validation AP@50-95, followed by a secondary refinement search around the best selected configuration from the first stage: RF-DETR-2XL with focus-stacked inputs. 

\textbf{Metrics.} We evaluate detection performance using precision, recall, and average precision (AP), following the COCO evaluation protocol. Precision and recall are reported at the confidence threshold that maximizes the $F_1$ score. 
A detection is considered correct if its predicted box and matched ground-truth box satisfy an IoU exceeding a specified threshold. We report AP@50 as well as AP@50-95, which penalizes coarse localization more strictly, both calculated using an IoU threshold of 0.5.

\subsection{Results}
\label{sec:results}

Table~\ref{tab:results} reports test-set performance for all four pairs of models and pre-processing conditions. RF-DETR-2XL outperforms YOLO26-L for both focal plane compression methods, by AP@50 of 0.053 and 0.017 for focus stacking and focal plane selection, respectively, with even larger margins in recall. The two architectures also respond differently to the choice of pre-processing method: YOLO26-L shows a considerable increase of 4\% in AP@50 from focus-stacked to selected focal plane images, a gap unlikely to be explained by training noise alone, while RF-DETR-2XL demonstrates similar performance for both focal plane compression methods. The overall best configuration, RF-DETR-2XL on focus-stacked images, achieves an AP@50 of 0.879 and AP@50-95 of 0.642, and is used as the final model for evaluating our I/O optimization methods.


\begin{table}[t]
\centering
\caption{Detection performance for all four pairs of model and focal plane compression configurations. Best result in every column is written in bold.}
\label{tab:results}
\footnotesize
\setlength{\tabcolsep}{3pt}
\begin{tabular}{@{}llcccc@{}}
\toprule
\textbf{Model} & \textbf{Pre-processing} & \textbf{AP@50} & \textbf{AP@50-95} & \textbf{Prec.} & \textbf{Rec.} \\
\midrule
YOLO26-L & Focus stacking & 0.826 & 0.594 & 0.825 & 0.752 \\
YOLO26-L & Focal plane selection & 0.860 & 0.604 & 0.838 & 0.757 \\
\midrule
RF-DETR-2XL & Focus stacking & \textbf{0.879} & \textbf{0.642} & \textbf{0.840} & \textbf{0.799} \\
RF-DETR-2XL & Focal plane selection & 0.877 & 0.637 & 0.833 & 0.792 \\
\bottomrule
\end{tabular}
\end{table}

\begin{figure}[t]
\centering
\setlength{\tabcolsep}{1pt}
\begin{tabular}{ccc}
\includegraphics[width=0.32\linewidth]{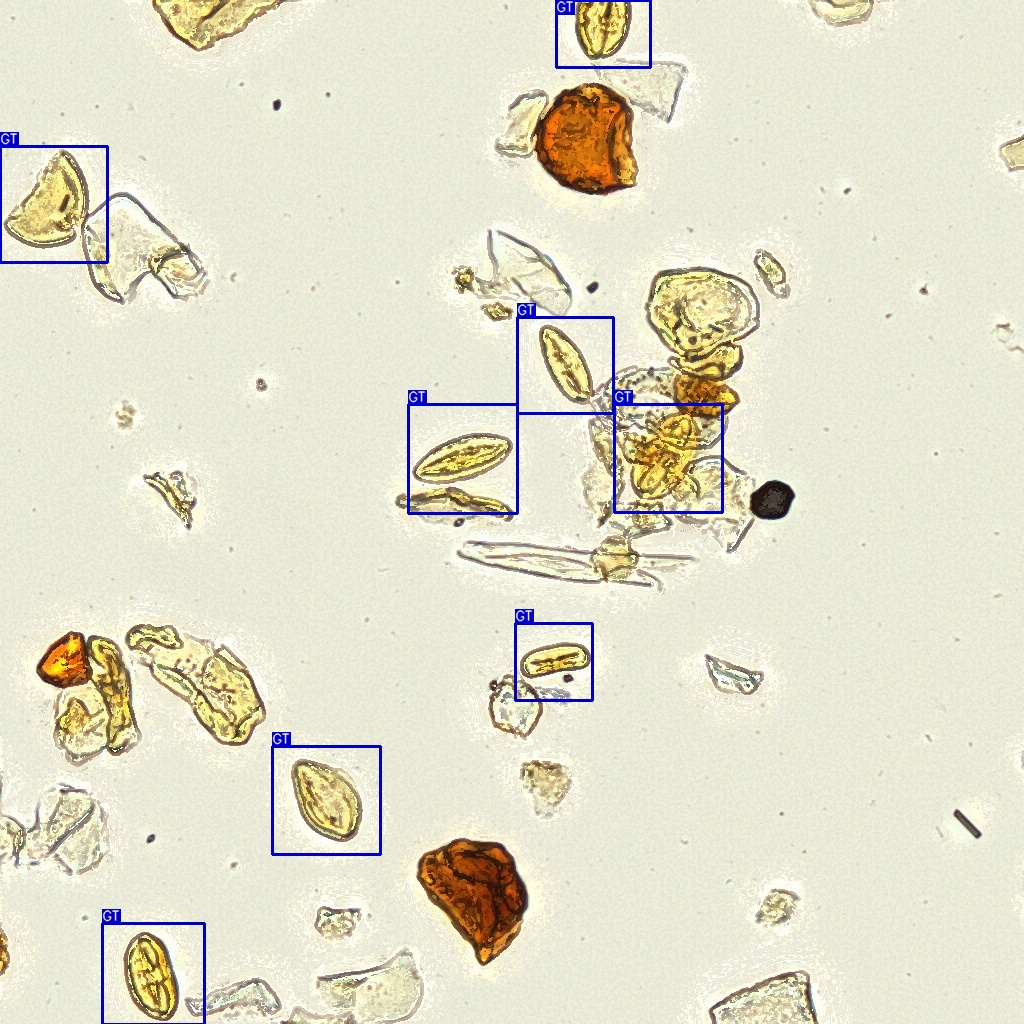} &
\includegraphics[width=0.32\linewidth]{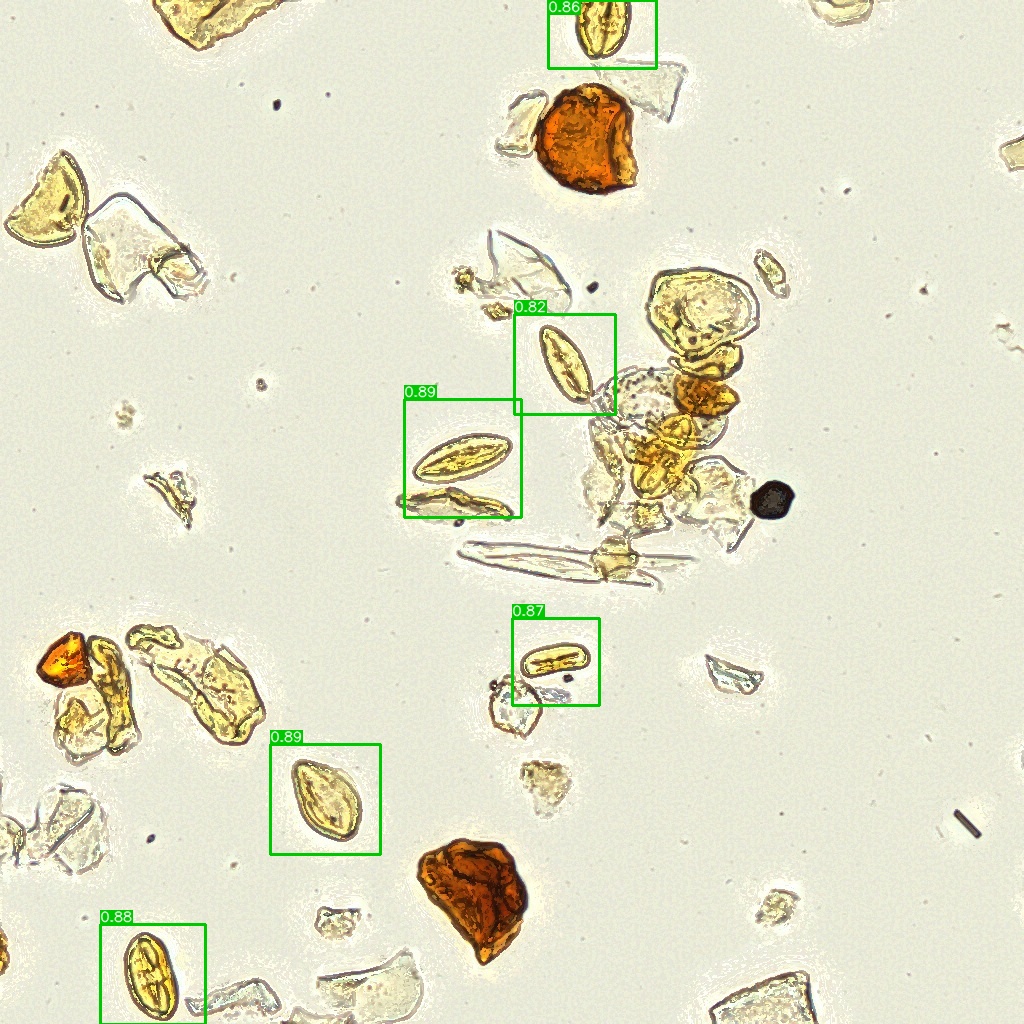} &
\includegraphics[width=0.32\linewidth]{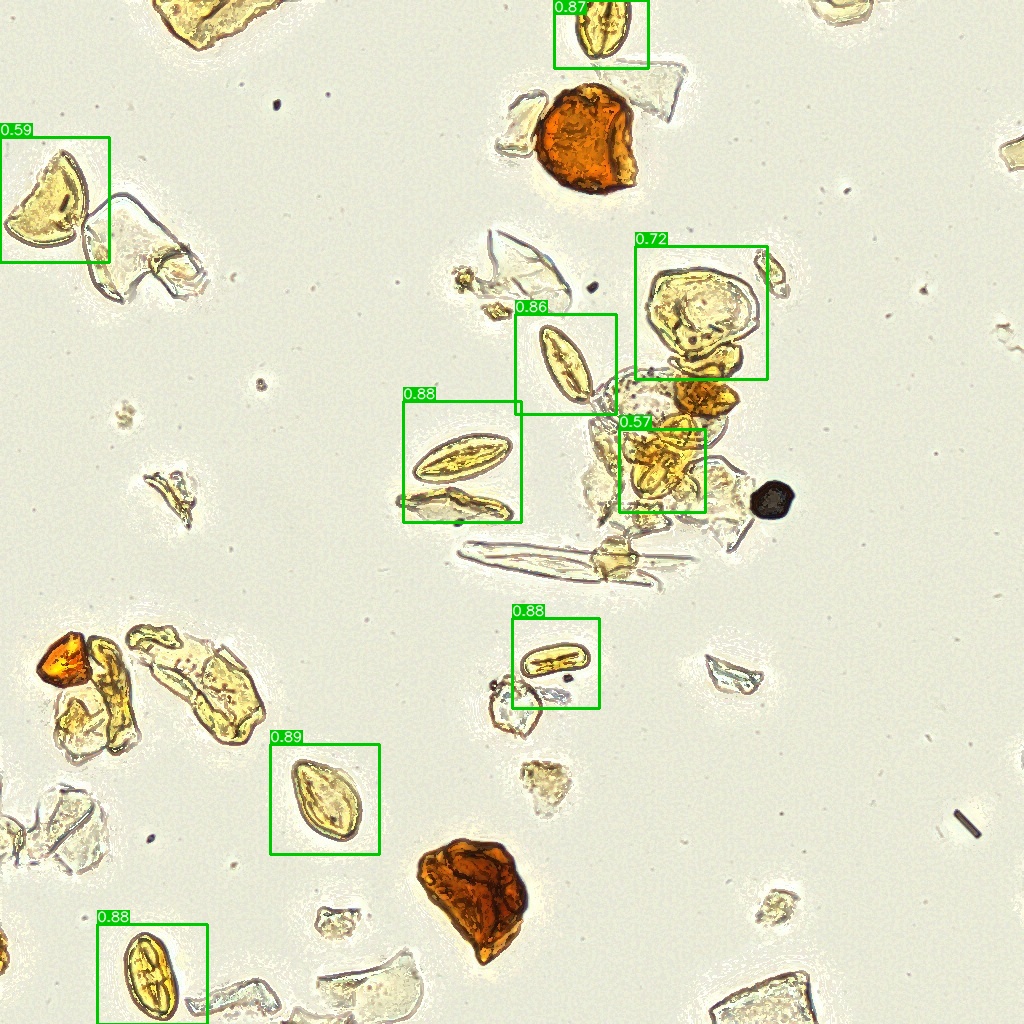} \\
\small (a) Ground truth & \small (b) YOLO26-L & \small (c) RF-DETR-2XL
\end{tabular}
\caption{Qualitative detection results on a representative held-out test tile (focus-stacked input). Blue boxes in (a) are ground-truth annotations; green boxes in (b) and (c) are model predictions.}
\label{fig:qualitative}
\end{figure}

Figure~\ref{fig:qualitative} shows tile-level predictions from both models alongside the ground truth annotations for a representative held-out test tile, both trained on focus-stacked images. RF-DETR-2XL recovers more true positives in visually dense, overlapping regions, consistent with its higher recall in Table~\ref{tab:results}, while YOLO26-L misses specimens more frequently in cluttered areas.

To evaluate the full end-to-end pipeline of Section~\ref{sec:whole-slide} at operational scale, we ran both the final RF-DETR-2XL and YOLO26-L models on a representative held-out test slide using focus-stacked tiles (Table~\ref{tab:timing}, rows 1--2). With the original read path, 
RF-DETR-2XL generates inference slower than YOLO26-L by roughly 50 minutes per slide. Since both models are intended for offline, whole-slide inference rather than real-time use, this overhead may be considered an acceptable cost for the higher accuracy of RF-DETR-2XL. However, these inference times far exceed the time required for the digitization of slides.

With the optimized read path of Section~\ref{sec:io}, whole-slide inference with RF-DETR-2XL drops to 51 minutes per slide in production (reading and focal compression 0.18--0.24\,s per tile and inference 0.15--0.20\,s per tile, at four CPU workers feeding a single GPU), roughly an order of magnitude faster end-to-end (Table~\ref{tab:timing}, row 3). At this rate, whole slide inference now keeps pace with digitization: whole-slide inference on a slide is obtained in under an hour on a single GPU, so newly scanned slides can be processed as they are produced rather than accumulating a backlog.


\begin{table}[t]
\centering
\caption{End-to-end whole-slide inference time (focus-stacked input) using a single NVIDIA L40S GPU. Rows 1--2: original read path, single held-out test slide. Row 3: optimized read path of Section~\ref{sec:io}; values are means over five production slides.}
\label{tab:timing}
\small
\setlength{\tabcolsep}{8pt}
\begin{tabular}{@{}llcc@{}}
\toprule
\textbf{Model} & \textbf{Read path} & \makecell{ \textbf{Whole-slide} \\ \textbf{Inference} } & \makecell{ \textbf{Tile} \\ \textbf{Inference} } \\
\midrule
YOLO26-L & Original & 7h 52m & 2.60s \\
RF-DETR-2XL & Original & 8h 42m & 2.75s \\
RF-DETR-2XL & Optimized & 51m & 0.41s \\
\bottomrule
\end{tabular}
\end{table}

Finally, deploying a model requires a single operating confidence threshold. We select $\tau^\star$ by maximizing the $\mathrm{F}_1$ score on the test set, which gives $\tau^\star = 0.404$ for the best overall configuration (RF-DETR-2XL, focus-stacked), with precision $0.824$, recall $0.810$, and $\mathrm{F}_1 = 0.817$ at that threshold. Since $\tau^\star$ is chosen on the same data used to report model performance, it represents a best-case operating point rather than an unbiased estimate of deployment performance. Lowering $\tau^\star$ trades precision for recall, favoring more comprehensive slide coverage at the cost of additional false positives.
\section{Discussion}
\label{sec:conclusion}

\subsection{Model Performance}

We achieve the best inference performance using the RF-DETR-2XL architecture, trained and evaluated on focus-stacked images across all metrics, including precision, recall, AP@50, and AP@50-95. The transformer-based RF-DETR model also outperformed our CNN baseline, YOLO26. This advantage likely arises from the architectural differences between the models. The transformer backbone of RF-DETR applies self-attention across the full feature map, enabling it to model long-range dependencies and global context more comprehensively than YOLO26. This approach may be particularly advantageous for palynomorph slides, which are characterized by high visual density and complex backgrounds, requiring contextual understanding to distinguish biological entities from inorganic debris.

Although RF-DETR achieved higher AP and recall than YOLO26, precision remained comparable between the two models. Qualitative results showed that RF-DETR substantially reduced false negatives at the cost of occasional false positives, particularly in areas of high occlusion. The higher AP indicates that these additional false positives were less detrimental than the comparatively larger number of missed ground-truth detections produced by YOLO. This difference may result from RF-DETR's matching strategy in training, which enforces one-to-one assignment between predictions and ground-truth objects, thereby reducing missed detections in dense or crowded regions where multiple nearby objects might otherwise compete for the same prediction.

For our particular application, the higher recall of RF-DETR is preferred. False positives can be eliminated through expert review, whereas false negatives represent undetected specimens. RF-DETR, therefore, better aligns with workflows requiring comprehensive specimen discovery. The optimal confidence threshold for RF-DETR indicates that the model generates well-calibrated, moderately high-confidence detections for a broad fraction of the specimen distribution. In a deployment context, researchers may choose to further lower or raise the threshold, depending on whether downstream review capacity or data completeness is the binding constraint.

\subsection{Comparison of Focal Plane Compressions Methods}
\label{sec:focus_vs_plane}

While the best focal plane selection substantially improved the performance of YOLO26 compared to focus stacking, it provided no discernible benefit for RF-DETR, which achieved near-identical results across both the focal plane compression strategies. Focus stacking can bring structures at different depths into focus within a single tile, but its pixel-wise selection introduces artifacts, such as halo edges, blending seams, and other high-frequency artifacts at depth discontinuities, that are absent in single-plane images. We attribute the performance differences of YOLO26 between the two focal plane compression methods to these artifacts. In contrast, RF-DETR demonstrated slight improvements using focus-stacked images. These results likely indicate that the larger representational capacity of RF-DETR was robust to the localized artifacts that focus stacking can introduce, allowing it to effectively leverage the sudden changes in sharpness that might be lost in a single representative focal plane. We also note that the aggressive data augmentation pipeline may have reduced much of the sharpness benefit that focus stacking provides, such that either model no longer significantly benefited from enhanced image contrast. However, our results generally suggest that while the optimal focal plane compression method may be dependent on the detector architecture, more robust models can better leverage sharper, focus-stacked images.

\subsection{Limitations and Future Work}

While the performance of RF-DETR establishes a promising baseline for automated palynology, there are several constraints that warrant further investigation. A relatively small sample size of 82 annotated slides may have resulted in a coarse train/validation/test split that might have lacked sufficient stratification of palynomorph subtypes. During development, we observed a consistent gap between validation and test performance, suggesting uneven difficulty of detection across splits, potentially driven by differences in geographic origin, staining protocols, and specimen density between slides. A more balanced, stratified sampling approach based on morphological complexity, slide quality, and/o palynomorph subtype distribution would likely yield a more reliable estimate of generalization performance in future work. 

Additionally, while both focus stacking and focal plane selection enable the use of standard 2D object detection architectures, they discard volumetric information -- through multi-image blending in focus stacking and through the loss of structural context from other planes in focal plane selection. Models specifically targeted toward multifocal object detection, capable of processing raw multifocal z-stacks, could potentially improve detection quality by retaining depth information end-to-end. However, we conjecture that the inference time of these models would be much higher compared to our efficient 2D approach.

Finally, our current focus on detection does not yet account for palynomorph subtype identification, which is a separate challenge due to the subtle taxonomic features required for differentiation, and could be an objective of future work.


\section{Conclusion}

Automated detection of palynomorphs in digital microscopy represents a critical step toward scaling palynological research. Manual annotation workflows are labor-intensive and constrained by the rapidly increasing volume of high-resolution slide data, limiting both throughput and reproducibility. In this work, we developed an end-to-end automated palynomorph detection pipeline for multifocal whole-slide images that integrates structured data handling, I/O optimization, depth-aware pre-processing, modern object detection architectures, and boundary-aware NMS to address these challenges. 

The proposed solution demonstrates that advanced detection models, particularly RF-DETR, can effectively capture complex visual patterns in palynomorphs, achieving an AP@50 of 0.879 and an AP@50-95 of 0.642. By incorporating multifocal information, the pipeline enhances the reliability of object detection in visually dense and heterogeneous images. Furthermore, we demonstrate the efficiency and scalability of the proposed pipeline, reducing the time for detection to under one hour. These improvements represent a meaningful step toward robust, automated analysis in this domain. Beyond performance gains, this work establishes a generalizable framework for large-scale microscopy analysis, especially for multifocal slide images. Our pipeline is designed to be extensible to other imaging-based scientific applications, reducing reliance on expert-driven labeling while maintaining reproducibility and scalability.

Overall, this study highlights the potential of modern AI and machine learning approaches to transform manual workflows in palynology and paleontology. By enabling data-driven analysis of palynomorph datasets, the proposed framework supports more comprehensive and efficient reconstruction of past environmental conditions and Earth's climate history.

{
    \small
    \bibliographystyle{ieeenat_fullname}
    \bibliography{main}
}

\end{document}